\documentclass[letterpaper, 10 pt, conference]{ieeeconf}  

\IEEEoverridecommandlockouts                              

\usepackage{amsmath} 
\usepackage{bbm}
\usepackage{dsfont}
\usepackage{xcolor}
\usepackage{amssymb}
\usepackage{graphicx}
\usepackage{mathtools}
\usepackage{amsmath}
\usepackage{gensymb}
\usepackage{float}
\usepackage{multirow}
\usepackage{makecell}
\usepackage{mathtools}
\usepackage{hyperref}
\hypersetup{
    colorlinks=true,
    linkcolor=blue,
    filecolor=magenta,      
    urlcolor=blue,
    pdftitle={Overleaf Example},
    pdfpagemode=FullScreen,
    }

\usepackage{tabularx}
    \newcolumntype{L}{>{\raggedright\arraybackslash}X}
\usepackage[utf8]{inputenc}
\usepackage[english]{babel}

\usepackage{amsthm}

\usepackage{subcaption}

\newcommand{\ours}{VLM-GroNav}

\title{\LARGE \bf VLM-GroNav: Robot Navigation Using Physically Grounded Vision-Language
Models in Outdoor Environments
}

\author{Mohamed Elnoor, Kasun Weerakoon, Gershom Seneviratne, Ruiqi Xian, Tianrui Guan, \\ Mohamed Khalid M Jaffar, Vignesh Rajagopal, and Dinesh Manocha \\
\small{Supplemental version including Tech Report and Video at \url{http://gamma.umd.edu/vlm-gronav/}}
\thanks{Authors are with University of Maryland, College Park. This work was supported in part by ARO Grant W911NF2310352 and Army Cooperative Agreement W911NF2120076.}
}

\begin{document}

\maketitle
\thispagestyle{empty}
\pagestyle{empty}

\begin{abstract}

We present a novel autonomous robot navigation algorithm for outdoor environments that is capable of handling 
diverse terrain traversability conditions. Our approach, \ours,  uses vision-language models (VLMs) and integrates them with physical grounding that is used to assess intrinsic terrain properties such as deformability and slipperiness. We use proprioceptive-based sensing, which provides direct measurements of these physical properties, and enhances the overall semantic understanding of the terrains.
Our formulation uses in-context learning to ground the VLM's semantic understanding with proprioceptive data to allow dynamic updates of traversability estimates based on the robot's real-time physical interactions with the environment. 
We use the updated traversability estimations to inform both the local and global planners for real-time trajectory replanning.
We validate our method on a legged robot (Ghost Vision 60) and a wheeled robot (Clearpath Husky), in diverse real-world outdoor environments with different deformable and slippery terrains.
In practice, we observe significant improvements over state-of-the-art methods by up to {50}\% increase in navigation success rate.

\end{abstract}

\section{Introduction} \label{sec:intro}

Autonomous robot navigation in outdoor environments has many challenges due to the variability and complexity of natural terrains \cite{wijayathunga2023challenges}. These terrains often exhibit unpredictable physical properties, such as deformability and slipperiness, which may vary considerably with changing weather conditions \cite{chen2024identifying,amco}. Traditional navigation methods rely mostly on vision-based sensors, such as cameras and LiDAR, to estimate terrain traversability \cite{guan2022ga,li2023seeing,frey2024roadrunner,agarwal2023legged}. While these sensors provide useful information about the environment, they can be insufficient or misleading in certain situations. For example, a terrain that appears solid and stable in a visual image might deform under the robot's weight and could lead to potential slippage or sinkage~\cite{morlando2023online,taghavifar2021novel}.

In robotic manipulation, haptic sensing has been used to enhance decision-making by physically grounding the robot's perception through measuring the physical properties of objects via tactile and proprioceptive-based sensing~\cite{lai2024vision,Guo2024PhyGrasp,gao2024physically}. Similarly, in robotic navigation, recent approaches have incorporated proprioceptive sensing information derived from the robot's internal sensors, such as joint encoders and force sensors, to design improved navigation methods~\cite{amco,fu2022coupling,karnan2024wait}. Proprioception offers direct, real-time measurements of the terrain’s physical characteristics at the robot's current location, which provides more insights that are not accessible through visual sensors. However, current proprioception methods typically lack the ability to predict the traversability of the terrain in the vicinity of the robot, thereby reducing their effectiveness in dynamic and complex environments~\cite{pronav}.

\begin{figure}[t]
      \centering  \includegraphics[width=0.9\columnwidth,height=5cm]{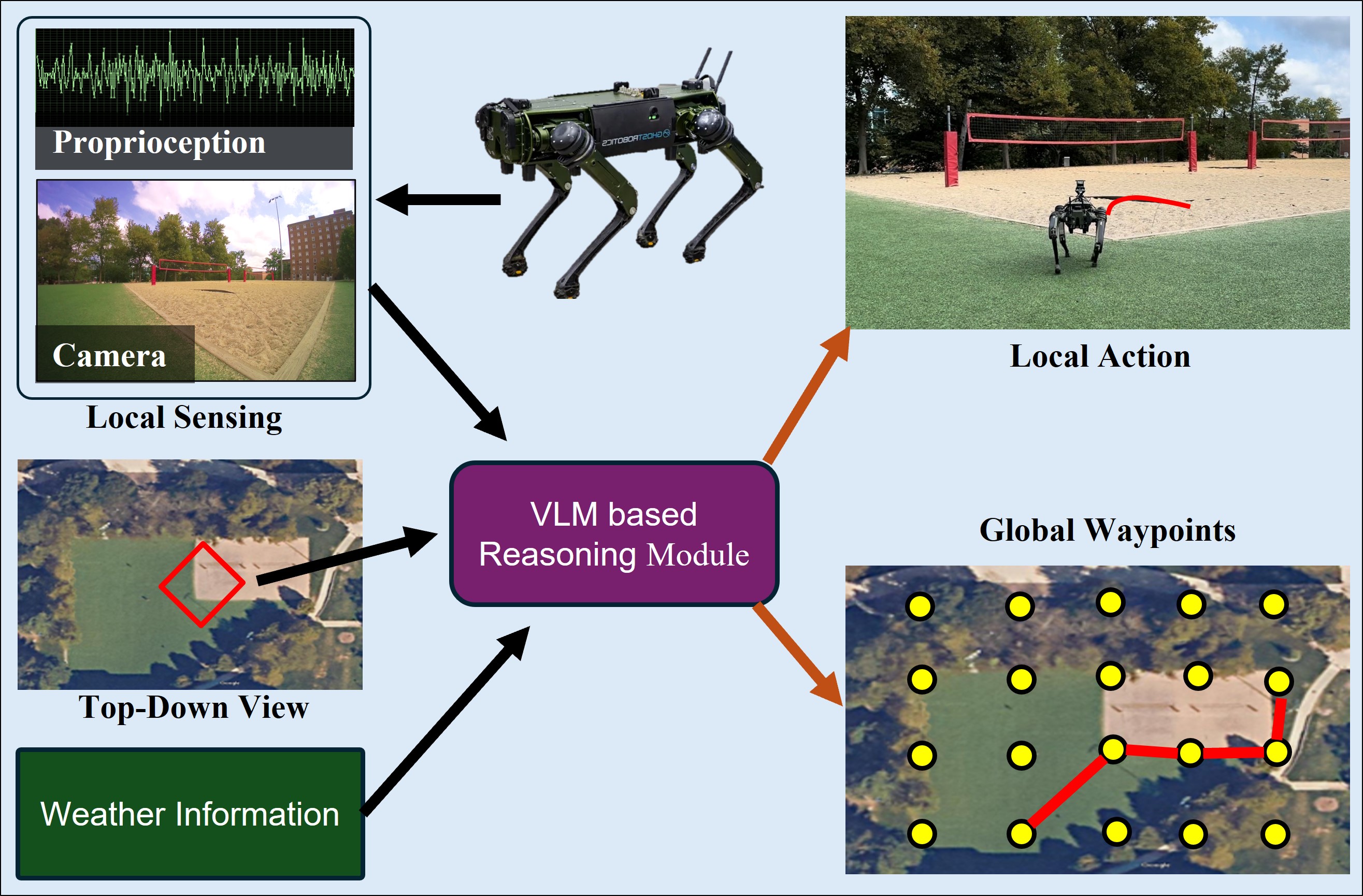}
      \caption {\small{Overview of our \ours \,\!\! system: Our method uses the given information to achieve a navigation objective. We leverage VLMs and aerial imagery to estimate initial terrain traversability. The robot's local exteroceptive and proprioceptive sensors guide the VLM to update the traversability and replan the robot's trajectory.} 
      }
      \label{fig:cover-image}
      \vspace{-20pt}
\end{figure}

Conversely, satellite or aerial imagery can offer a broader, top-down perspective of the environment. Such overhead point of views can help identify potential challenges or obstacles in advance~\cite{аksamentov2020approach}. 
However, using such imagery for effective terrain analysis presents additional challenges; these images may 
lack sufficient detail to capture the complex characteristics of natural terrain~\cite{balaska2021enhancing}. Furthermore, existing datasets for aerial image segmentation largely focus on urban environments~\cite{boguszewski2021landcover,shen2021s2looking,van2021multi} and may not be directly applicable to outdoor terrains with varying physical properties (e.g., sand, grass, mud, snow). In such cases, vision-language models (VLMs) can be useful due to their broad training on internet-scale data, which allows them to infer prior estimates about terrain traversability from visual cues in satellite imagery~\cite{al2023vision}.

VLMs have demonstrated remarkable success in tasks such as zero-shot and few-shot classification \cite{zhao2024vlm}, semantic understanding \cite{ha2022semantic}, and common-sense reasoning \cite{kwon2024toward}. Their ability to reason about the environment using both visual and textual inputs presents promising opportunities for advancing robotic navigation \cite{yokoyama2024vlfm}. VLMs excel at interpreting visual and language data, yet these models may not take into account critical physical properties of the terrain such as deformability, surface hardness, and friction that are essential for navigating complex outdoor environments.



On the other hand, tactile and proprioception-based sensing have been widely explored in different robotic applications~\cite{li2020review}, particularly to assess physical properties like material rigidity or fragility~\cite{qu2023recent}. However, only few past studies have investigated its usage as \textit{a modality} to ground or guide large language models' reasoning~\cite{yu2024octopi}. 




\textbf{Main contributions:}
We present \ours, a novel navigation method that integrates Vision-Language Models (VLMs) with proprioception-based sensing. This approach enables VLMs to ground their decision-making in the robot's physical interactions with the environment and improve both local and global path planning. Our key contributions are as follows:

\begin{itemize}
    \item \textbf{\textit{A physically informed terrain traversability estimation module:}} We incorporate proprioception-based sensing using in-context learning to enhance VLM predictions of terrain traversability, focusing on physical properties such as deformability (for legged robots) and slipperiness (for wheeled robots). By providing proprioceptive inputs to the VLM through in-context learning \cite{dong2024surveyincontextlearning}, we enable the model to understand how these properties influence navigation, resulting in more accurate and reliable estimates compared to zero-shot predictions. This dynamic fine-tuning of the VLM’s understanding of diverse terrains significantly improves both global and local planning, particularly in complex or unfamiliar environments.
    
    

    \item \textbf{\textit{An novel adaptive high-level global planner:}} We introduce a global planner that leverages a large VLM that's informed by the terrain’s estimated traversability, to guide waypoint selection and trajectory planning. Our approach visually marks the aerial imagery to provide the VLM with spatial context and terrain information, enhancing its understanding of the environment. The marked image guides the VLM to select optimal waypoints based on the navigation objective, such as minimizing trajectory length or avoiding hazardous regions like slippery or deformable terrains. This process allows for dynamic trajectory re-planning, informed by both visual cues and updated traversability estimates.

    \item \textbf{\textit{A real-time adaptive local planner:}} We introduce a frontier-based local planner that dynamically adjusts the robot's trajectory by integrating proprioceptive feedback with a compact Vision-Language Model (VLM). This planner performs zero-shot terrain classification to identify candidate frontiers in the robot's immediate field of view, assigning terrain types to each potential path. By incorporating a new frontier cost term into the Dynamic Window Approach \cite{DWA} objective function, our method prioritizes trajectories toward more traversable terrains. This integration allows the robot to adapt its path in real-time based on terrain assessments.

    \end{itemize}
    
    Extensive experimental validation shows that our approach, \ours \,\! performs better than state-of-the-art methods by 50\% in terms of navigation success rate, 
     which highlights the benefits of integrating physically grounded information into VLM-based navigation.
     
    

\section{Related Work} \label{sec:related_work}

\subsection{Traversability Estimation}
Traversability estimation in outdoor environments relies on both exterioceptive and proprioceptive sensing. Exterioceptive approaches, such as those using vision-based sensors like cameras \cite{frey2024roadrunner,agarwal2023legged,julius2022scene} and LiDAR \cite{li2023seeing,sathyamoorthy2024mim}, are effective for creating terrain maps and categorizing surfaces through semantic segmentation \cite{guan2022ga}. However, their performance can be compromised by environmental factors like lighting changes and occlusions.
Proprioception, on the other hand, provides direct feedback on terrain interaction, crucial for navigating complex terrains \cite{pronav,karnan2024wait}. It has been used to adapt robot locomotion based on real-time terrain feedback, as seen in multi-modal sensor fusion techniques that combine proprioceptive data with visual and inertial inputs \cite{fu2022coupling}. While powerful, proprioception alone cannot predict terrain features ahead of the robot without additional sensor data \cite{amco}. Our approach integrates this real-time feedback with Vision-Language Models (VLMs) to adjust navigation trajectories dynamically based on changing terrain conditions.

\subsection{Local \& Global Path Planning}
Local planning focuses on immediate path adjustments to navigate obstacles and maintain smooth motion \cite{zhao2023intelligent}. Traditional methods often use local cost maps derived from exteroceptive sensors \cite{weerakoon2022terp,overbye2021path,wellhausen2023artplanner}, but these can be limited in terms of handling unpredictable and complex outdoor environments. Some approaches have incorporated proprioceptive data to enhance path planning, particularly in scenarios where visual data alone is insufficient \cite{amco}. Our approach builds on these advancements by using VLMs to generate context-aware local trajectories that consider both immediate obstacles and the robot's physical capabilities.

Conversely, global planning methods broadly fall into two categories: search-based and optimization-based approaches. Search-based methods, such as A* or Dijkstra’s, rely on heuristic-driven searches to find an optimal path from the robot’s current location to a distant goal, typically minimizing objectives like travel distance or time \cite{ju2020path,luo2020surface,guo2020global}. Optimization-based approaches, on the other hand, frame the planning problem as an optimization task that considers multiple constraints, such as minimizing energy consumption or avoiding obstacles \cite{samadi2013global,hong2024safe}. While these methods are effective in structured environments, they often require significant expert tuning to adapt to varying constraints in more complex settings, such as outdoor terrains.

In contrast, our approach leverages the generality and general reasoning capabilities of Vision-Language Models (VLMs) to achieve a broader range of objectives beyond simply minimizing path length. By integrating VLMs with global planning, our approach can consider additional factors such as avoiding deformable terrains, minimizing energy consumption, and adhering to traversability preferences, such as avoiding grass or traversing stable ground. This flexibility allows the robot to adjust its global path and waypoint selection dynamically based on the current environmental context.

\subsection{Foundation and Large Models in Navigation}

The integration of Foundation Models, Large Language Models (LLMs) and Vision-Language Models (VLMs) into robotics has marked a significant advancement in robot navigation \cite{sridhar2024nomad,zhu2024visual,shek2023lancar}, enabling robots to leverage their common-sense reasoning and visual understanding capabilities for complex decision-making tasks \cite{yokoyama2024vlfm}. These models have demonstrated their benefits across a range of robotic applications, from high-level task planning \cite{singh2023progprompt} to low-level control \cite{xu2024drivegpt4}.
One example, SayCan \cite{brohan2023can} integrates LLMs for high-level task planning, where it effectively converts natural language instructions into actionable robot behaviors. Similarly, LM-Nav \cite{shah2023lm} combines the strengths of GPT-3 and CLIP to navigate outdoor environments, using natural language instructions grounded in visual cues to guide the robot’s path.

VLMs have also shown promise in bridging the gap between visual perception and language understanding in robotics. Visual Language Maps \cite{10160969} proposes a spatial map representation that fuses vision-language features with a 3D map. ViNT \cite{shah2023vint} introduces a foundation model for visual navigation that leverages a transformer-based architecture to generalize across diverse environments and robot embodiments. Recently, NoMaD \cite{sridhar2024nomad} has taken a unified approach by employing a diffusion-based policy that enables both task-agnostic exploration and goal-directed navigation in unseen environments.
While significant progress has been made in using LLMs and VLMs for high-level navigation goals like reaching specific locations or following instructions \cite{sathyamoorthy2024convoi}, challenges remain in applying these models to physically grounded, real-time navigation, particularly in complex outdoor terrains with varying physical terrain properties (e.g. deformability). 
Our work bridges this gap by combining VLMs with proprioception-based to adapt navigation trajectories based on real-time terrain conditions.

\section{Background}\label{sec:background}
In this section, we describe our system setup, state our assumptions, and introduce
key concepts used in our work.

\subsection{Setup, Assumptions, and Conventions}
Our formulation assumes a robot equipped with an RGB camera, 3D LiDAR, IMU, and GPS, with a common coordinate frame centered at the robot’s center of mass. The $X$, $Y$, and $Z$ axes point forward, left, and up, respectively. The camera provides RGB images $I_{\text{RGB}, t}$, and the LiDAR captures surrounding 3D point clouds for obstacle detection. Additionally, GPS offers global localization for the robot, while the IMU provides orientation and motion feedback.
Both, legged and wheeled robots make use of the same controller architecture, receiving linear and angular velocity commands $(v, \omega)$ in the robot’s frame. As for proprioceptive sensing, in legged robots, we use joint encoders to measure $p_{\text{joint}, i}$ (position), $v_{\text{joint}, i}$ (velocity), and $f_{\text{joint}, i}$ (force). For wheeled robots, we use wheel odometry for state estimation.


For navigation, the robots utilize both global and local planners. The global planner leverages aerial imagery and GPS to generate high-level global waypoints, while the local planner uses real-time sensory feedback, including proprioception to adjust the robot's trajectory based on terrain conditions.

\subsection{Motion Planner}\label{motion planner}

We adopt the Dynamic Window Approach \cite{DWA}, which is a widely used real-time local planner in mobile robots, designed to ensure collision-free and dynamically feasible navigation. DWA computes admissible velocities $(v, \omega)$ by considering the robot’s kinematic limits and obstacle proximity. The search space for these velocities is defined as:
\begin{equation}
\mathcal{V}_r = \{(v, \omega) \mid v \in [0, v_{\max}], \omega \in [-\omega_{\max}, \omega_{\max}] \}
,
\end{equation}

where $v$ and $\omega$ are the linear and angular velocities, and $v_{\max}$, $\omega_{\max}$ are the maximum allowable velocities.
At each time step, candidate trajectories are propagated over a fixed horizon $\Delta T$, and each trajectory is evaluated by minimizing an objective function:
\begin{equation}
J(v, \omega) = \rho_{1} \cdot \text{head}(v, \omega) + \rho_{2} \cdot \text{dist}(v, \omega) + \rho_{3} \cdot \text{vel}(v, \omega).
\end{equation}

The terms in the objective function are defined as follows: $\text{head}(v, \omega)$ measures alignment with the goal, $\text{dist}(v, \omega)$ ensures obstacle avoidance, and $\text{vel}(v, \omega)$ encourages faster movement. $\rho_{1}$, $\rho_{2}$, $\rho_{3}$ are weighting factors. 

\begin{figure*}[t]
    \centering
    \includegraphics[width=1.0\linewidth]{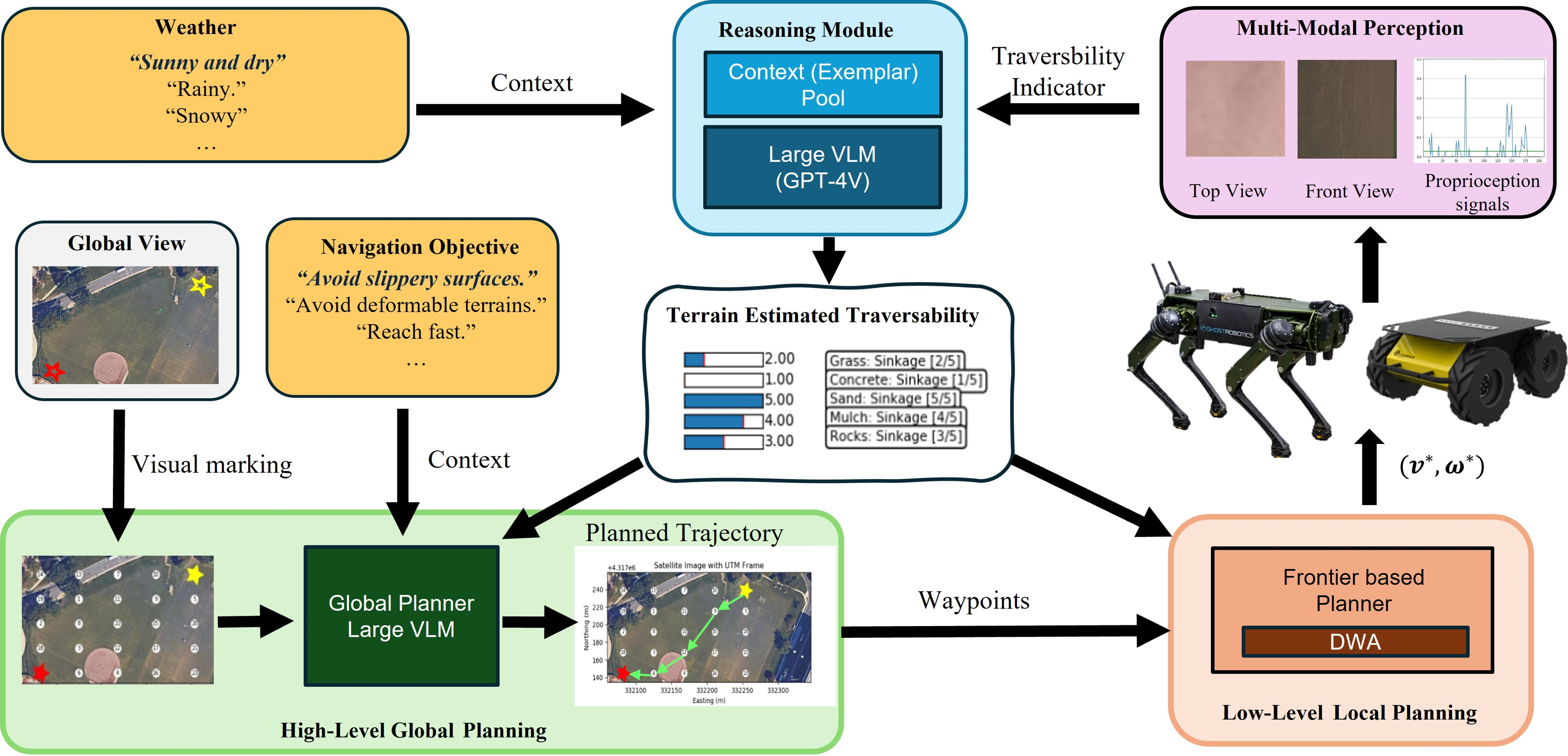}
    \caption{\small{The \ours \,\! system employs a reasoning module that integrates visual inputs from aerial imagery, weather conditions, and proprioceptive data through a large VLM to refine terrain traversability estimates and inform navigation decisions. The global planner uses these refined estimates to generate optimal waypoints visually marked on the aerial image to guide the robot toward the goal while achieving the goal objective. The local planner utilizes a frontier-based approach, incorporating real-time proprioceptive feedback to adapt the trajectory dynamically for both legged and wheeled robots for robust navigation.}}
    \label{fig:system-arch}
    \vspace{-10pt}
\end{figure*}

\section{Our Approach} 


We propose a novel navigation method that integrates Vision-Language Models (VLMs) with proprioceptive sensing to enable adaptive and robust navigation across complex outdoor terrains. 
The overall architecture of our method is shown in Fig \ref{fig:system-arch}. The core components of our method are:
(a) \textit{Traversability Estimation using Proprioception-based sensing}; 
(b) \textit{Physically Grounded Reasoning Module};
(c) \textit{High-Level Global Planner};
(d) \textit{Adaptive Local Planner}. 

\subsection{Traversability Estimation using Proprioceptive sensing}
We estimate terrain traversability using proprioceptive-based sensors. For legged robots, we calculate the robot's leg sinkage through the forces applied to the joints, which provide a direct measure of how the terrain deforms. For wheeled robots, we evaluate the terrain slipperiness.


\subsubsection{Deformability for legged robots}

We estimate terrain deformability for legged robots using the sum of squared forces applied to the joints, calculated as \( S_{\text{sinkage}} = \sum_{i=1}^{n} f_{\text{joint}, i}^2 \), where \( f_{\text{joint}, i} \) is the force experienced by the \(i\)-th joint, and \(n\) is the total number of joints. The sinkage indicator \(S_{\text{sinkage}}\) varies across different terrains. 
To compute the traversability indicator \(\tau\), we normalize the sinkage indicator by applying a scaling factor \(\Gamma\). The traversability indicator is then calculated as:

\begin{equation}
\tau_{\text{sinkage}} = \Gamma \cdot \frac{S_{\text{sinkage}} - S_{\text{min}}}{S_{\text{max}} - S_{\text{min}}},
\end{equation}

where \(S_{\text{min}}\) corresponds to the least deformable terrain (e.g. concrete), and \(S_{\text{max}}\) corresponds to the most deformable terrain (e.g. loose sand). 


\subsubsection{Slippage for wheeled robots}

For wheeled robots, we estimate slippage by comparing wheel odometry with LiDAR-based odometry as implemented in \cite{sathyamoorthy2022terrapn}. The difference between these measurements reflects the degree of slippage experienced by the robot. We define the slippage traversability indicator $\tau_{\text{slip}}$ as:
\begin{equation}
\tau_{\text{slip}} = \beta_1 (\Delta d_{\text{lidar}} - \Delta d_{\text{odom}}) + \beta_2 (\Delta \theta_{\text{lidar}} - \Delta \theta_{\text{odom}}),
\end{equation}
where $\Delta d$ and $\Delta \theta$ represent the distance traveled and the change in orientation over a time interval, obtained from LiDAR odometry and wheel odometry (odom), and $\beta_1$ and $\beta_2$ are weighting factors.

\subsection{Physically Grounded Reasoning Module}
The reasoning module combines visual and proprioceptive data to continuously update terrain traversability estimates and navigation strategies. It leverages VLMs to process visual inputs (aerial imagery and front camera views), and integrates real-time feedback from the robot's local sensors.

At the start of navigation, our autonomy stack queries a large VLM to classify terrain types $\mathcal{L} = \{\ell_1, \ell_2, \dots, \ell_n\}$ based on aerial imagery and weather data. Each terrain class is linked to parameters like the traversability indicator $\tau$, 
which are updated as the robot collects proprioceptive data.
During navigation, the robot captures $5m \times 5m$ patches from the front camera and aerial imagery. The traversability indicator ($\tau_{sinkage}$ and $\tau_{slip}$) are time-shifted to match the visual inputs, $\tau_{\text{shifted}}(t) = \tau(t - \Delta t)$.
Then we build the exemplar pool $\mathcal{E}_{\text{pool}}$, which includes aerial imagery, front camera views, aligned traversability indicators, and terrain classes. For a given terrain class $i$ we can denote:
\begin{equation}
\mathcal{E}_{\text{pool}}(i) = \{ (I_{\text{aerial}}(i), I_{\text{front}}(i), \tau_{\text{shifted}}(i), \ell_i) \}.
\end{equation}

We use in-context learning \cite{dong2024surveyincontextlearning} to refine terrain traversability and navigation cost estimates. By grounding VLM predictions with real-time proprioceptive feedback, the system adapts its understanding of the terrain as the robot interacts with it.
The VLM uses the exemplars and a text prompt to estimate the terrain's traversability as follows:

\begin{equation}
\tau_{\text{estimate}} = \text{VLM}(\mathcal{T}_{\text{prompt}}, \mathcal{E}_{\text{pool}}).
\end{equation}


\begin{figure*}[t]
    \centering
    \includegraphics[width=1.0\linewidth]{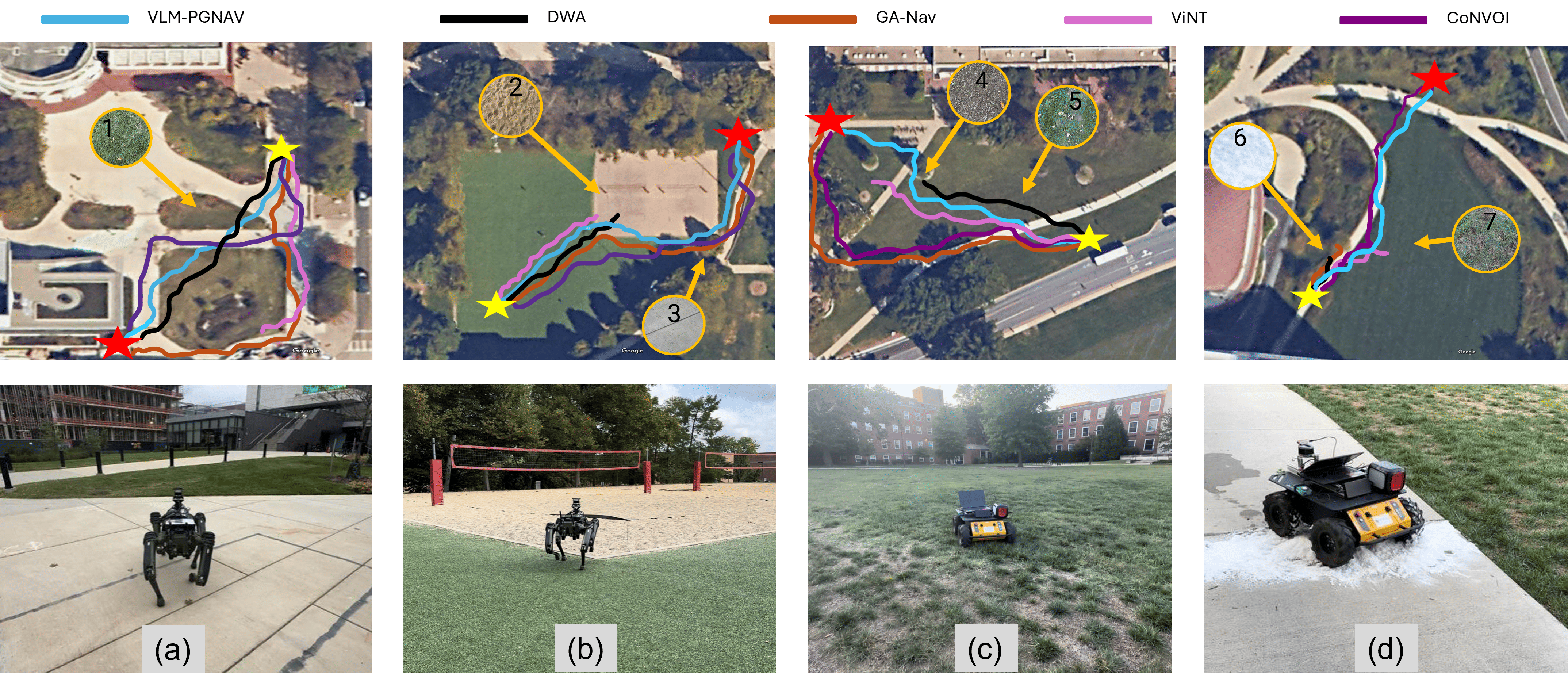}
    \caption{\small{Comparison of navigation trajectories across various environments using different methods: DWA (Black), GA-Nav (orange), CoNVOI (Dark purple), ViNT (light purple),  and our method \ours \,\! (sky blue). Yellow stat represents the start location and red star represents the goal location. (a) shows Scenario 1, (b) shows Scenario 2, (c) shows Scenario 3, and (d) shows Scenario 4. The top row shows the top-down image of the scene. Top images also contain circles with terrain pictures (1: Grass, 2: Sand, 3: Concrete, 4: Mulch, 5:Muddy Grass, 6:Snow, 7:Muddy Grass). Our Method demonstrates a more direct path with minimal detours. 
    }}
    \label{fig:expirments}
    \vspace{-10pt}
\end{figure*}

\subsection{High-Level Global Planner}

Our global planner uses aerial imagery and VLM to generate an optimal set of waypoints guiding the robot from its current position to the goal. 
 To enhance the VLM's ability to identify navigable regions, we apply visual marking to the aerial image. This involves overlaying numbered markers on potential waypoints within the image, which helps the VLM better understand the spatial layout and possible paths.
 The VLM is then prompted with this marked image and a navigation objective $T_{\text{objective}}$ to select the optimal sequence of waypoints that lead to the goal while adhering to specific navigation objectives.

The sequence of waypoints selected by the VLM, denoted as $\mathcal{W}_{\text{global}} = \{ w_1, w_2, \dots, w_k \}$ form the initial global path. As the robot navigates, traversability estimates may change due to new proprioceptive feedback from the reasoning module. When this occurs, the global planner replans by re-querying the VLM with updated terrain information:

\begin{equation}
\mathcal{W}_{\text{new}} = \text{VLM}(\mathcal{T}_{\text{objective}}, \mathcal{I}_{\text{marked}}, \tau_{\text{estimate}}  ),
\end{equation}

where $I_{\text{marked}}$ is the aerial image after applying the visual marking.
The updated waypoints $W_{\text{new}}$ are then passed to the local planner. The local planner adjusts the robot's trajectory in real-time based on local terrain conditions and sensory feedback.

\subsection{Adaptive Local Planner}

Our local planner adapts in real-time to changes in terrain traversability by integrating proprioceptive feedback with a light VLM (with low inference time). 
We employ a CLIP \cite{radford2021learning} for zero-shot terrain classification to identify candidate frontiers to the left, middle, and right sides in front of the robot. These frontiers denoted as $P = \{ p_L, p_C, p_R \}$, are projected to the image frame and visually marked in the robot's RGB camera image and passed to the CLIP for zero-shot terrain classification. Each waypoint is assigned a terrain type $\ell_i$.

To integrate terrain traversability into the planning process, we introduce a new cost term, the \emph{frontier cost}, into the DWA's objective function. This term prioritizes trajectories leading toward more traversable frontiers. The modified objective function $G(v, \omega)$ is:


\begin{equation}
G(v, \omega) = J(v,\omega) + \rho_{4} \cdot \phi(v, \omega),
\end{equation}

where, $J(v,\omega)$ is the objective function defined in Section \ref{motion planner}. And the \emph{frontier cost} term $\phi(v, \omega)$ is calculated as, 
\begin{equation}
\phi(v, \omega) = \min_{p \in P} \left( d(\eta(v, \omega), p) \cdot \tau_{estimate}(p) \right).
\end{equation}

Here, $\eta(v, \omega)$ represents the trajectory resulting from linear and angular velocities $v$ and $\omega$, $d(\eta(v, \omega), p)$ is the Euclidean distance between the endpoint of the trajectory $\eta(v, \omega)$ and the frontier point $p$, and $\tau_{estimate}(p)$ 
is the traversability estimate assigned to frontier $p$ by the reasoning module.


\section{Results and Analysis}

\subsection{Implementation}

For the real-world experiments, we utilize both the Ghost Vision 60 legged robot and the Clearpath Husky wheeled robot. The Ghost Vision 60 is equipped with a front-facing wide-angle camera, an OS1-32 LiDAR, GPS, and an onboard Intel NUC 11 system, which includes an Intel i7 CPU and an NVIDIA RTX 2060 GPU. The Clearpath Husky is equipped with a Velodyne VLP16 LiDAR, a Realsense D435i camera, GPS and a laptop containing Intel i9 processor and Nvidia RTX2080 GPU. We use GPT-4o API for the reasoning and global planning modules and CLIP-based \cite{radford2021learning} zero-shot terrain classification in the local planner.

\subsection{Comparison Methods}

\begin{itemize}

    \item DWA \cite{DWA}: A baseline
    motion planner that performs simple collision avoidance
    and goal-reaching behaviors.

    \item GA-Nav \cite{guan2022ga}: A method that uses semantic segmentation-based terrain understanding to generate traversability costs for navigation.



    \item CoNVOI \cite{sathyamoorthy2024convoi}: A method that uses Vision Language Models (VLMs) to generate context-aware trajectories by identifying the robot’s environment and following implicit and explicit navigation behaviors in both indoor and outdoor settings.

    \item ViNT \cite{shah2023vint}: A general-purpose foundation model for visual navigation that uses a Transformer-based architecture trained on diverse robotic navigation datasets to adapt across different tasks and robot embodiments.


\end{itemize}

\subsection{Evaluation Metrics}

\begin{itemize}

    \item \textbf{Success Rate}: The ratio of successful navigation trials where the robot was able to reach its goal without freezing or colliding with obstacles.


    \item \textbf{Normalized Trajectory Length:} The ratio between the robot’s trajectory length and the straight-line distance to the goal averaged across both successful and unsuccessful trials.

    \item \textbf{IMU Energy Density}: The total aggregated squared acceleration values measured by the IMU sensors across the x, y, and z axes, calculated over the successful trials \cite{try2023vibration}: \( E_i = \sum_{n=1}^{N} a_{i,n}^2 \), and \( E_{\text{Total}} = E_{ax} + E_{ay} + E_{az} \), where \( a_i \) represents one of the three acceleration signals (x, y, and z axes), and \( N \) is the number of IMU readings along the trajectory.
    


\end{itemize}

\begin{table}[t]
\centering
\resizebox{\columnwidth}{!}{%
\begin{tabular}{ |c|c|c|c|c|c| } 
\hline
\textbf{Scenario} & \textbf{Method} & \makecell{\textbf{Success}\\ \textbf{Rate} $\uparrow$} & \makecell{\textbf{Norm.}\\ \textbf{Traj. Length}} & \makecell{\textbf{IMU Energy}\\ \textbf{Density} $\downarrow$} \\ 
\hline
\multirow{6}{*}{\rotatebox[origin=c]{0}{\textbf{Scen. 1}}} 
& DWA \cite{DWA} & 30 & 0.98 & 32934 \\
& GA-NAV \cite{guan2022ga} & 60 & 1.31 & 28419 \\
& CoNVOI \cite{sathyamoorthy2024convoi} & 80 & 1.23 & 26391 \\
& ViNT \cite{shah2023vint}  & 90 & 1.16 & 21234 \\
& \ours\ w/o GP & 90 & 1.01 & 22347 \\
& \ours\ w/o ICL & 90 & 1.12 & 21293 \\
& \ours\ (ours) & \textbf{100} & 1.07 & \textbf{20753} \\
\hline
\multirow{6}{*}{\rotatebox[origin=c]{0}{\textbf{Scen. 2}}} 
& DWA \cite{DWA} & 10 & 1.06 & 28052 \\
& GA-NAV \cite{guan2022ga} & 50 & 1.34 & 19823 \\
& CoNVOI \cite{sathyamoorthy2024convoi} & 60 & 1.25 & 18572 \\
& ViNT \cite{shah2023vint} & 40 & 1.17 & 27822 \\
& \ours\ w/o GP & 80 & 1.32 & 15239 \\
& \ours\ w/o ICL & 60 & 1.65 & 19234 \\
& \ours\ (ours) & \textbf{100} & 1.23 & \textbf{16273} \\
\hline
\multirow{6}{*}{\rotatebox[origin=c]{0}{\textbf{Scen. 3}}} 
& DWA \cite{DWA} & 10 & 1.25 & 26392 \\
& GA-NAV \cite{guan2022ga} & 40 & 1.48 & 21394 \\
& CoNVOI \cite{sathyamoorthy2024convoi} & 60 & 1.23 & 22342 \\
& ViNT \cite{shah2023vint} & 30 & 1.21 & 30208 \\
& \ours\ w/o GP & 70 & 1.09 & 23217 \\
& \ours\ w/o ICL & 60 & 1.31 & 25310 \\
& \ours\ (ours) & \textbf{90} & 1.02 & \textbf{21982} \\
\hline
\multirow{6}{*}{\rotatebox[origin=c]{0}{\textbf{Scen. 4}}} 
& DWA \cite{DWA} & 20 & 0.97 & 31923 \\
& GA-NAV \cite{guan2022ga} & 40 & 1.51 & 28345 \\
& CoNVOI \cite{sathyamoorthy2024convoi} & 60 & 1.44 & 29473 \\
& ViNT \cite{shah2023vint} & 40 & 1.16 & 25451 \\
& \ours\ w/o GP & 60 & 1.13 & 24341 \\
& \ours\ w/o ICL & 50 & 1.40 & 23310 \\
& \ours\ (ours) & \textbf{90} & 1.01 & \textbf{21049} \\
\hline
\end{tabular}%
}
\caption{\small{Performance comparisons across four scenarios using different navigation methods over $10$ trials. The table shows success rates, normalized trajectory lengths, and IMU energy density. Our method achieves the highest success rate in all scenarios, along with lower IMU energy density. All metrics are averaged over both the successful and unsuccessful trails (reaching the goal).}}
\label{tab:comp-results}
\end{table}

\subsection{Testing Scenarios}

\begin{itemize}
    \item \textbf{Scenario 1:} \textit{Dry Grass, Muddy Grass, Concrete}: The legged robot transitions from stable concrete to slippery, deformable muddy grass. 
    \item \textbf{Scenario 2:} \textit{Dry Grass, Sand, Concrete}: This considers legged robot on loose sand with varying deformability.
    \item \textbf{Scenario 3:} \textit{Concrete, Dry Grass, Muddy Grass}: Wheeled robot transitions from firm surfaces (concrete) to deformable terrains (muddy grass).
    \item \textbf{Scenario 4:} \textit{Concrete, Snow, Muddy Grass}: Wheeled robot navigates slippery surface (snow) over concrete.
\end{itemize}

\subsection{Analysis and Comparison}

We evaluate the performance of our method both qualitatively and quantitatively and compare it with other navigation methods. Table \ref{tab:comp-results}
provides a summary of the results across four scenarios while the qualitatively results are shown in Fig \ref{fig:expirments}. \ours\ consistently achieves the highest success rates in all scenarios and demonstrates its capability to adapt to diverse terrains. Also, \ours\ results
in lower vibrations in both robots across all scenarios whencompared to other methods; evidenced by the IMU energy density. 

Scenarios 1 and 2 involve the legged robot navigating multiple terrains with different deformability. \textbf{In Scenario 1}, our method first traverses the dry grass region and adapts its trajectory in the second region due to its proprioceptive sensing of the deformability of the muddy grass. While DWA takes a short path, it neglects the muddy grass region which causes an increase in the energy density and reduced success rate. VLM-based (CoNVOI/ViNT) and segmentation (GA-Nav) methods prefer to take the concrete path, a longer detour to the goal. \textbf{In Scenario 2}, where there is grass and sand, DWA and ViNT struggle to navigate the sand part with the loose sand and cause a rise in vibration. GA-Nav and CoNVOI take a long path and GA-Nav struggles in the border due to changes in vibration, consequently causing motion blur. Whereas, our method adapts its trajectory to traverse the sandy region due to its updated traversability from the sinkage detection. This change increased the path slightly but resulted in low vibration.


Scenarios 3 and 4 involve the wheeled robot navigating through unstructured and slippery terrains, \ours\ excels at maintaining a high success rate and reduced IMU energy density. 
\textbf{In Scenario 3}, our approach records a lower normalized trajectory length compared to GA-Nav and CoNVOI because they preferred to take the concrete path. ViNT navigated through the grass but it didn't update its grass traversability and frequently failed in the muddy grass. \textbf{In Scenario 4}, snow presented challenges to DWA and GA-Nav due to its slipperiness which resulted in a lower success rate and caused an increase in energy density (increased instability). CoNVOI and ViNT struggled in the muddy region of the grass. Our method, \ours \,\! was able to consistently update its traversability estimation of the concrete with snow and navigated through a more stable path. 

\textbf{Ablation}: We noticed that removing the global planning part affected the navigation performance (success rate), particularly in Scenario 3. We believe that happens because the scenario contains multiple terrains with different physical properties. Furthermore, we tested our method without the In-context learning (ICL) part, which makes the reasoning module rely on the zero-shot capabilities of the VLM. We observe that this results in errors in predicting the terrain's traversbility while navigating, which in turn ill-informs the local and global planners, causing failures. That is evident in Scenarios 2 \& 3. 

Overall, our method shows superior performance in both, instability reduction and path selection compared to other methods. It achieves that by leveraging visual and proprioceptive data to dynamically update traversability estimates. This results in more stable navigation and efficient energy usage, which is critical for autonomous outdoor robots.

\section{Conclusions, Limitations and Future Work}

We introduce \ours, a novel method that integrates Vision-Language Models (VLMs) with proprioceptive sensing to enhance terrain traversability estimation for both legged and wheeled robots. 
Our method uses the obtained traversability estimates to inform both the local and global planners for real-time trajectory re-planning.
However, our method has some limitations. \mbox{\ours}\,\! relies on accurate GPS for global localization, which can be limiting in GPS-denied environments. Additionally, the method currently lacks the integration of other sensing modalities, such as thermal or hyperspectral cameras, which could improve navigation in low-visibility conditions.
In future work, we aim to incorporate these alternative sensor modalities to enhance terrain assessment capabilities and explore methods to improve localization without GPS. Additionally, optimizing VLM processing will be crucial for faster decision-making in dynamic and complex environments.

\bibliographystyle{IEEEtran}
\bibliography{References}

\end{document}